# A Multi-model Combination Approach for Probabilistic Wind Power Forecasting

You Lin, Ming Yang, *Member, IEEE*, Can Wan, *Member, IEEE*,
Jianhui Wang, *Senior Member, IEEE*, and Yonghua Song, *Fellow, IEEE*

*Abstract*—Short-term probabilistic wind power forecasting can provide critical quantified uncertainty information of wind generation for power system operation and control. As the complicated characteristics of wind power prediction error, it would be difficult to develop a universal forecasting model dominating over other alternative models. Therefore, a novel multi-model combination (MMC) approach for short-term probabilistic wind generation forecasting is proposed in this paper to exploit the advantages of different forecasting models. The proposed approach can combine different forecasting models those provide different kinds of probability density functions to improve the probabilistic forecast accuracy. Three probabilistic forecasting models based on the sparse Bayesian learning, kernel density estimation and beta distribution fitting are used to form the combined model. The parameters of the MMC model are solved based on Bayesian framework. Numerical tests illustrate the effectiveness of the proposed MMC approach.

## I. INTRODUCTION

As one of the most cost-effective renewable power sources, wind power generation has been widely applied in modern power systems. According to the statistics launched by the Global Wind Energy Council, till the end of 2015, the global installed capacity of wind power has reached 432,883 MW. Meanwhile, in several countries, the wind power penetration has already achieved a relatively high level, e.g., 42% electricity consumption of Demark was provided by wind power in 2015. However, because of the inherent intermittency and uncertainty, large-scale wind power integration brings serious challenges to power system operation and control [1], [2].

To accommodate the variability of wind power, probabilistic wind generation forecasting techniques are rapidly developed in the last decade. Compared with the deterministic wind generation forecasting, probabilistic forecasting can provide quantified uncertainty involved in wind power forecasting and then benefit power system reserve setting, unit commitment, market trading, and so on [3],[4],[5],[6],[7],[8]. To improve the forecast accuracy, both parametric and non-parametric models have been proposed for probabilistic forecasting of wind power, such as the quantile regression [9],[10], ensemble approach [11], the adaptive resampling approach [12], radial basis function neural network [13], bootstrapped extreme learning machine [14], conditional kernel density estimation [15], direct interval forecasting [16], hybrid neural network [17], sparse Bayesian learning (SBL) approach [18], and classification approach [19].

Most of the typical probabilistic forecasting approaches are based on an individual forecast model. However, it would be difficult to find an individual forecast model which is perfect for all kinds of wind farms, especially for probabilistic models based specific probability distribution assumption, since the characteristics of wind generation may change significantly from one wind farm to another and the complicated statistical nature of wind power prediction error. Previous studies have found that the combined forecast mean is more skillful than each member model [20]. Therefore, designing a combined model which can exploit the advantages of different kinds of forecast models is quite desirable. Actually, it has been well recognized that the combined model can provide better forecast results than a single model [21]. Several combined models have already been established for the point forecasts of wind generation [21]. The ensemble probabilistic forecasting method was adopted to predict the meteorological data, including wind speed and precipitation, based on Bayesian model averaging (BMA) [22]. The individual distributions of BMA are in accordance with the same distribution type. However, the above probabilistic combination approaches are either based on the forecast error statistical distribution of the combined point forecast approach or the combination of the same member distributions.

In this paper, a multi-model combination (MMC) approach is

This work was supported in part by National High-Tech Research and Development Program (863 Program) of China (2014AA051901), National Basic Research Program (973 Program) of China (2013CB228205), The Fundamental Research Funds of Shandong University (2015JC028), International S&T Cooperation Program of China (2014DFG62670), National Science Foundation of China (51477091), and Hong Kong RGC Theme Based Research Scheme (T23-407/13-N). J. Wang's work was supported by the U.S. Department of Energy (DOE)'s Office of Electricity Delivery and Energy Reliability.

Y. Lin is with Key Laboratory of Power System Intelligent Dispatch and Control, Shandong University, Jinan, Shandong 250061 China, and also with College of Electrical Engineering, Zhejiang University, Hangzhou 310027, China.

M. Yang is with Key Laboratory of Power System Intelligent Dispatch and Control, Shandong University, Jinan, Shandong 250061 China, and also with Argonne National Laboratory, Argonne, IL 60439 USA.

C. Wan is Department of Electrical Engineering, Tsinghua University, Beijing 100084, China, and also with Argonne National Laboratory, Argonne, IL 60439 USA.

J. Wang is with the Energy Systems Division, Argonne National Laboratory, Argonne, IL 60439 USA (e-mail: jianhui.wang@anl.gov).

Y. Song is with Department of Electrical Engineering, Tsinghua University, Beijing 100084, China, and also with College of Electrical Engineering, Zhejiang University, Hangzhou 310027, China (e-mail: yhsong@tsinghua.edu.cn).

proposed for probabilistic wind generation forecasting by extending the conventional BMA. It combines multiple probabilistic forecasting models those provide different types of probability density functions (PDFs) to achieve better prediction results through combining the superiorities of different individual distribution forecasting methods. The weights of member distributions reflect their contributions to the forecasts, which are solved adaptively to realize the best performance for different wind farms. Three different kinds of mature probabilistic wind power forecasting models based on sparse Bayesian learning (SBL), kernel density estimation (KDE) and beta distribution estimation (BDE) are adopted to form the combined forecasting model. The parameters of the MMC model are firstly solved via expectation maximizing (EM) algorithm. Through the EM algorithm, the MMC model can realize a better forecast expectation than the individual models. Then parameters are further optimized by maximizing the performance of the forecast distribution to get a more calibrated probabilistic forecast results. The effectiveness of the proposed MMC approach has been verified by the simulation experiments on the dataset of Global Energy Forecasting Competition 2014 (GEFCOM 2014).

## II. Combined Model for Probabilistic Forecasting

### A. Prediction Uncertainty

Standard regression models typically execute forecasts conditionally on the selected model. However, a deterministic forecasting model can hardly be perfect on some specified criterion because the prediction error may always exist, even for a carefully designed model. The prediction error cannot be avoided because of the following reasons:

Firstly, the forecasting model is an abstraction, simplification and interpretation of reality. The incompleteness of the model and the mismatch between the model and the real causal structure of a system always result in prediction errors.

Secondly, data uncertainties will also cause the forecasting errors. The unsuitable or unreliable training data may cause misleading parameters of the forecasting model. Meanwhile, the measurement or estimation errors of the input variables, e.g., weather forecasting errors, will also be propagated into the final wind power forecasting.

Moreover, the nonlinear, non-stationary natures of wind power and the complex structures of the weather system and power plant system will also make the design of a perfect forecasting model impossible.

Therefore, probabilistic forecasting becomes very meaningful to estimate wind power prediction uncertainties. More specifically, we can take parametric probabilistic wind generation forecasting as an example here. Denote wind power at time $t$ by $y_t$ and the input vector of the prediction model by $x_t$. The training data set can be expressed as

$$D = \{x_t, y_t\}_{t=1}^{T}, \quad (1)$$

where $T$ represents the size of the data set.

The forecasting target can be formulated as

$$y_t = g(x_t) + \varepsilon(x_t), \quad (2)$$

where $g(x_t)$ denotes the true regression, and $\varepsilon(x_t)$ denotes the noise.

Prediction error is produced by model misspecification and data uncertainty. Approximating the true regression $g(x_t)$ by $\hat{g}(x_t)$, the prediction error can be expressed by

$$y_t - \hat{g}(x_t) = [g(x_t) - \hat{g}(x_t)] + \varepsilon(x_t), \quad (3)$$

where $y_t - \hat{g}(x_t)$ denotes the prediction error, and $g(x_t) - \hat{g}(x_t)$ is the approximation error of the true regression.

For conventional parametric probabilistic forecasting techniques, the noise is commonly supposed to be normally distributed with zero mean and variance $\hat{\sigma}_\varepsilon^2$ associated with input variables [23]. Assuming the model uncertainty and data uncertainty in (3) are statistically independent, the variance of prediction uncertainty, expressed by $\hat{\sigma}_y^2(x_t)$, can be derived by

$$\hat{\sigma}_y^2(x_t) = \hat{\sigma}_g^2(x_t) + \hat{\sigma}_\varepsilon^2(x_t). \quad (4)$$

The parametric predictive distribution based on normal model can be expressed as

$$p(y_t | \hat{g}(x_t)) \sim N(\hat{g}(x_t), \hat{\sigma}_y^2(x_t)). \quad (5)$$

However, as the complex statistical characteristics of wind power prediction error, the prediction uncertainty cannot be precisely described on basis of the parametric probability distribution model, such as normal distribution.

### B. Basics of Bayesian Model Averaging

As mentioned above, a perfect forecasting model can hardly be obtained, and thus, how to eliminate or reduce the bias of probabilistic wind power forecasting can be an eternal topic. Actually, instead of designing a more sophisticated forecasting model, model combination is a feasible way in statistics to improve the preciseness of probabilistic forecasting [24]. As early as 1960s, it has been illustrated that combing forecasts from different persons is beneficial [20], which is confirmed by the later studies [25], [26]. And it is further verified that this principle is valid not only for the performance of subjective forecasters but also for the objective multi-model forecasting systems [27]. The variations in physics and numerics of the forecasting models make the solution more reliable [28].

Bayesian model averaging aims to obtain the "best" forecast model conditioning on the combination of several individual forecast models. Given the training dataset $y_{Tr}$ the objective $y$ to be predicted on the basis of $K$ individual models $F_1, F_2, …, F_K$, the law of BMA provides the combined forecast PDF, described as

$$p(y_t|F_1,\cdots,F_K) = \sum_{k=1}^{K} p(y_t|F_k)p(F_k|y_{Tr}), \quad (6)$$

where $p(y_t|F_1,\cdots,F_K)$ is the predictive PDF of $y_t$ obtained by the combined model, $F_k$ is the $k$th member model, $p(y_t|F_k)$ is the predictive PDF generated by $F_k$, and $p(F_k|y_{Tr})$ is the posterior probability of model $F_k$ indicating the contribution of $F_k$ to the forecast satisfying

$$0 \leq p(F_k|y_{Tr}) \leq 1, \quad (7)$$

$$\sum_{k=1}^{K} p(F_k|y_{Tr}) = 1. \quad (8)$$

Thus the BMA predictive PDF is a weighted average of the conditional PDFs based on the individual models. The BMA approach has been adopted in weather forecasts, such as wind speed and temperature. However, the conditional PDFs given by the individual model are in accordance with the same distribution type, for example, normal distribution in temperature forecasts. In this paper, the MMC approach is a more flexible model by extending the BMA member models into multiple distribution types.

## III. MULTI-MODEL COMBINATION APPROACH

### A. Formulations of MMC

To exploit the advantages of different probabilistic forecasting models, an MMC approach is developed for probabilistic forecasting of wind power generation.

Based on the principles of model combination, a general model of MMC with $K$ member models can be expressed as

$$p(y_t|F_1,\cdots,F_K) = \sum_{k=1}^{K} w_k p_k(y_t|F_k), \quad (9)$$

where $w_k$ is posterior probability, as well as the weight of model $F_k$ satisfying

$$0 \leq w_k \leq 1, \quad (10)$$

$$\sum_{k=1}^{K} w_k = 1. \quad (11)$$

It should be noted that $F_1,\cdots,F_K$ can be any kinds of forecasting models those are able to provide predictive PDF results. Moreover, the explanatory variables for the member models need not to be the same. Here, we denote $F_k$ as a general model with expectation $\bar{y}_{k,t}$ and standard deviation $\sigma_k$ whose predictive PDF can be described by

$$p_k(y_t|F_k) = p(\bar{y}_{k,t}, \sigma_k^2). \quad (12)$$

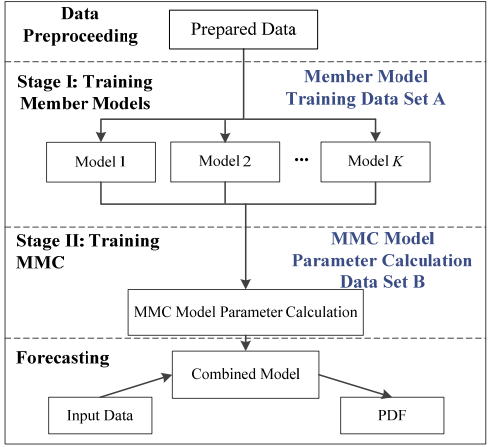

Fig. 1 Framework of MMC.

The overall framework of MMC is briefly described in Fig. 1. The training of the MMC model is divided into two main stages, i.e., the member model training stage and the MMC parameters estimation stage. Accordingly, the historical data are also divided into two data sets. With the first data set, the member models are trained according to their own principles. Then the parameters of the MMC model are estimated in the second stage by maximum likelihood and the EM algorithm on the basis of the optimization dataset. Furthermore, the parameters achieved from the EM algorithm are further optimized to realize its optimal distribution forecast effect.

### B. Member Models

In this paper, three different probabilistic forecasting models, i.e., SBL, KDE, and BDE, are adopted to form the combined model. SBL is a nonlinear sparse parametric forecasting model, which can provide conditional Gaussian PDF of wind generation [18]. KDE is a nonparametric probabilistic forecasting model whose output needs not obey any predetermined distribution type. In this paper, the adopted KDE model is also conditional in order to capture the variation of the predictive PDF with explanatory variables [15]. BDE is a statistical approach for PDF estimation, which can estimate the PDF of wind generation according to the Beta distribution [29]. It can be found that the selected member models have quite different mechanisms, and their performance on probabilistic forecasting of wind generation will also be very different. As to be illustrated in cases studies, these distinct member models can indeed get improvement from each other. The details of the adopted member models are introduced as follows.

*1) Sparse Bayesian Learning*

SBL is a sparse kernel-based probabilistic forecasting model which can be expressed as

$$y_t = \boldsymbol{\omega}^T \boldsymbol{\phi}(\boldsymbol{x}_t) + \varepsilon_t = \sum_{i=1}^{M} \omega_i K_i(\boldsymbol{x}_t, \boldsymbol{x}_i) + \omega_0 + \varepsilon_t, \quad (13)$$

where $\boldsymbol{x}_t$ and $y_t$ are the input vector and output of the model, $M$ is the number of kernels, $\boldsymbol{\omega}=[\omega_0, \omega_1,\cdots,\omega_M]^T$ is the weight vector, $K_i(\boldsymbol{x}_t, \boldsymbol{x}_i)$ is a Gaussian kernel function, $\boldsymbol{\phi}(\boldsymbol{x}_t)=[1, K_1(\boldsymbol{x}_t,\boldsymbol{x}_1),\cdots,K_M(\boldsymbol{x}_t,\boldsymbol{x}_M)]^T$, and $\varepsilon_t$ is the Gaussian residual with zero mean and variance $\sigma_\varepsilon^2$.

To avoid overlearning, SBL estimates the parameters by using Bayesian rules. More specifically, in (15), the weights are treated as Gaussian variables $N(0,\alpha_i^{-1}), i=0,1,\cdots,M$, where $\alpha_i$, $i=0,1,\ldots,M$ are hyper-parameters. And then, according to the Bayesian rules, the covariance matrix and mean value vector of $\boldsymbol{\omega}$ can be obtained with $N$ groups of samples $\{(\boldsymbol{x}_n, y_n)\}_{n=1}^{N}$, as

$$\boldsymbol{\Sigma} = \left(\sigma_\varepsilon^{-2}\boldsymbol{\Phi}^T\boldsymbol{\Phi} + \boldsymbol{A}\right)^{-1}, \quad (14)$$

$$\boldsymbol{\mu} = \sigma_\varepsilon^{-2}\boldsymbol{\Sigma}\boldsymbol{\Phi}^T\boldsymbol{y}, \quad (15)$$

where $\boldsymbol{\Phi}=[\boldsymbol{\phi}(\boldsymbol{x}_1),\boldsymbol{\phi}(\boldsymbol{x}_2),\cdots,\boldsymbol{\phi}(\boldsymbol{x}_M)]^T$, $\boldsymbol{y}=[y_1,y_2,\cdots,y_M]$, and $\boldsymbol{A}=diag(\alpha_0,\alpha_1,\cdots,\alpha_M)$. Parameters $\alpha_i$, $i=0,1,\ldots,M$ and $\sigma_\varepsilon^2$ in (14) and (15) are estimated by using maximum likelihood estimation.

Finally, for each new input $\boldsymbol{x}_t^*$, the distribution of $y_t^*$ can be obtained according to (13) as $N(\hat{y}_t, \hat{\sigma}_t^2)$, expressed as,

$$\hat{y}_t = \boldsymbol{\mu}^T \boldsymbol{\phi}(\boldsymbol{x}_t^*), \quad (16)$$

$$\hat{\sigma}_t^2 = \sigma_\varepsilon^2 + \boldsymbol{\phi}\left(\boldsymbol{x}_t^*\right)^T \boldsymbol{\Sigma} \boldsymbol{\phi}\left(\boldsymbol{x}_t^*\right). \tag{17}$$

*2) Kernel Density Estimation*

KDE is a non-parametric PDF estimation approach which estimates the PDF of the target random variable without pre-assuming its distribution type. Therefore, KDE is believed to have more adaptability than the parametric estimation approaches.

With the data set of historical observations $\{(\boldsymbol{x}_m, y_m)\}_{m=1}^M$, the conditional PDF of the target value $y_t^*$ corresponding to the new input $\boldsymbol{x}_t^*$ can be estimated according to the law of total probability, expressed as

$$p(y_t^* | \boldsymbol{X} = \boldsymbol{x}_t^*) = \frac{f_{XY}(\boldsymbol{x}_t^*, y_t^*)}{f_X(\boldsymbol{x}_t^*)} = \frac{1}{h_Y} \sum_{i=1}^M \omega(\boldsymbol{x}_t^*, \boldsymbol{x}_i) K\left(\frac{y_t^* - y_i}{h_Y}\right), \tag{18}$$

where $f_{XY}(\cdot)$ is the joint PDF of $X$ and $Y$, $f_X(\cdot)$ is the joint PDF of $X$, $K(\cdot)$ is a Gaussian kernel function, $h_Y$ is the bandwidth parameter of $Y$ that controls the smoothness of the estimated PDF, and $\omega(\boldsymbol{x}_*, \boldsymbol{x}_m), m = 1, 2, \ldots, M$ are the weights of the kernels which can be expressed as

$$\omega(\boldsymbol{x}_t^*, \boldsymbol{x}_i) = \frac{K(\mathbf{H}_X^{-1}(\boldsymbol{x}_t^* - \boldsymbol{x}_i))}{\sum_{j=1}^M K(\mathbf{H}_X^{-1}(\boldsymbol{x}_t^* - \boldsymbol{x}_j))}, \tag{19}$$

where $\mathbf{H}_X = diag(h_1, h_2, \cdots, h_D)$ is a bandwidth matrix which controls the smoothness of the explanation variable vector $X$, and $D$ is the dimension of $X$.

In KDE, the bandwidth is an important exogenous parameter which is determined by using Silverman's rule of thumb in this paper [30].

*3) Beta Distribution Estimation*

Normalized wind generation should be within the interval [0, 1]. Therefore, the beta distribution bounded by 0 and 1 is widely applied to describe the prediction uncertainty of wind generation [29].

The PDF of beta distribution $Beta(\alpha, \beta)$ can be described as

$$p(y_t; \alpha, \beta) = \frac{y_t^{\alpha-1}(1-y_t)^{\beta-1}}{B(\alpha, \beta)}, \tag{20}$$

where $\alpha$ and $\beta$ are the shape parameters satisfying $\alpha, \beta > 0$, and $B(\alpha, \beta)$ is the Beta function.

The relation between the expected value $\mu$ and the variance $\sigma^2$ of Beta distribution and the shape parameters of Beta distribution can be expressed as

$$\alpha = \mu(\mu - \mu^2 - \sigma^2)/\sigma^2, \tag{21}$$

$$\beta = (1-\mu)(\mu - \mu^2 - \sigma^2)/\sigma^2. \tag{22}$$

Therefore, the shape parameters as well as the PDF can be obtained by estimating the expected value and variance from the samples.

*C. Parameter Estimation*

Parameters of MMC model include the parameters of the individual forecast model and the weights of the member models. The SBL and KDE models can be trained on basis of the training dataset independently, as described in the above member models. For BDE model, the shape parameters $\alpha$ and $\beta$ can be achieved by estimating its expectation and variance. In this paper, the expectation $\mu$ and variance $\sigma^2$ of BDE is estimated separately. The expectation $\mu$ can be estimated using any spot forecast models based on the training dataset. Here, support vector machine (SVM) is adopted to estimate the expectation $\mu$. The detailed estimation process of SVM is described in [19].

The parameters including weights of member models $w_1, \cdots, w_K$ and the variance $\sigma^2$ of BDE can be estimated by maximum likelihood on basis of the training data. The objective is to find the parameters those maximize the likelihood function, which is equal to maximize the log-likelihood function, defined as

$$l(w_1, \cdots, w_K, \sigma^2) = \sum_{n=1}^N \log\left(\sum_{k=1}^K w_k p_k(y_n | F_k)\right). \tag{23}$$

It is difficult to maximize the log-likelihood function analytically or numerically using nonlinear maximization approaches. Thus expectation-maximization (EM) algorithm is adopted to identify the maximum likelihood estimator, i.e., the parameters maximizing the likelihood function.

The EM algorithm is an iterative method used for finding the maximum likelihood estimators. The estimation process of EM algorithm can be described as follows:

a) Initialize the parameters to be estimated $\boldsymbol{\theta}^{(j)}(j=0)$.

b) Expectation (E) step. Estimating the expectation of the log-likelihood evaluated with the current parameters $\boldsymbol{\theta}^{(j)}$ using the following function

$$z_{n,k}^{(j)} = w_k p_k(y_n | F_k, \sigma^{(j-1)}) \Big/ \sum_{k=1}^K w_k p_k(y_n | F_k, \sigma^{(j-1)}). \tag{24}$$

c) Maximization (M) step. Update parameters maximizing the expected log-likelihood on basis of the E step.

$$w_k^{(j)} = \frac{1}{N} \sum_{n=1}^N z_{n,k}^{(j-1)}, \tag{25}$$

$$\sigma^{2(j)} = \frac{1}{N} \sum_{n=1}^N \sum_{k=1}^K z_{n,k}^{(j-1)} (y_n - \bar{y}_{n,k})^2. \tag{26}$$

where $\bar{y}_{n,k}$ is the predictive expectation of wind power by model $F_k$.

d) If the log-likelihood converges to a maximum likelihood, i.e., changes of the parameter values are no greater than the pre-set tolerances, stop the iteration and output the maximum likelihood estimator. Otherwise, go back to E step.

As a remedy of that the maximum likelihood estimators calculated by the EM algorithm may converge to local optimal ones, the estimate of parameters is further refined to optimize the distribution accuracy. Among the performance assessment criteria of probabilistic forecasting, continuous ranked probability score (CRPS) [31] is a comprehensive criterion that can assess the calibration and sharpness of the forecasted PDF simultaneously. The smaller the CRPS value is, the better the distribution forecast will be. The objective of the optimization is to minimize the CRPS value of the training data over a range of values of $w_1, \cdots, w_K$ and $\sigma^2$, centered at the EM estimation. And particle swarm optimization algorithm is adopted to find the optimal parameters.

## IV. CASE STUDIES

### A. Data Description and Pretreatment

MMC is tested on wind power data of Global Energy Forecasting Competition 2014 to verify its effectiveness [32]. The dataset includes wind power generation data and weather prediction results for 5 wind farms, given as zonal and meridional components. The wind forecasts are given at two heights, i.e., 10m and 100m above the ground level. The data range from Jan. 1, 2012 to Dec. 31, 2012, and the time resolution is one hour.

In the study, the PDFs of wind generation are predicted for future 24 hours. For the $t$-hour-ahead ($t$ = 1, 2, … 24) forecasting, the dataset is divided into two training sets (A and B) and a validation set, as shown in Table I. The training set A is used to train the member models while the training set B is used to optimize the weights of the member models. The validation set is used to test the performance of the obtained MMC model.

TABLE I. THE DATASET DESCRIPTION

| Data set | Description | Number |
|---|---|---|
| Training set A | 1/1/2012 1:00 a.m. - 1/5/2012 0:00 a.m. | $M$=2904 |
| Training set B | 1/5/2012 1:00 a.m. - 1/6/2012 0:00 a.m. | $N$=744 |
| Validation set | 1/6/2012 1:00 a.m. - 1/7/2012 0:00 a.m. | 720 |

The magnitude of the wind vector is the dominant contributory factor of wind generation. Therefore, the zonal and meridional components of wind speed are first transformed to the magnitude and angle of wind vector by (27) and (28), whose mechanism can be illustrated by Fig. 2.

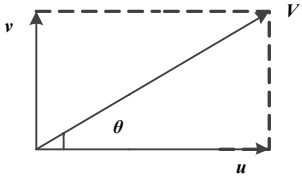

Fig. 2 Coordinate transform relations.

$$V = \sqrt{u^2 + v^2}, \quad (27)$$

$$\theta = \begin{cases} \arctan(v/u), & u \geq 0, v \geq 0, \\ 2\pi + \arctan(v/u), & u \geq 0, v < 0, \\ \pi + \arctan(v/u), & u < 0, \end{cases} \quad (28)$$

where $u$ and $v$ are the zonal and meridional components of the wind vector respectively, $V$ is the magnitude of the wind vector (wind speed), and $\theta$ is the angle of the wind vector (wind angle).

Correlation coefficients can address the degree of relationship between two mathematical variables. To determine the input data of the forecast model, the autocorrelation of wind generation data is first tested. Fig. 3 shows the autocorrelation function of the observed wind generation series. It illustrates that wind generation has a high autocorrelation within 24 hours lags.

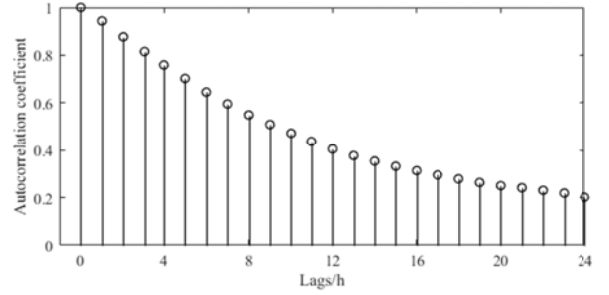

Fig.3 Autocorrelation coefficients of wind power.

Then, the cross-correlation between wind generation and wind speed/angle is also tested and listed in Table II. The results demonstrate that wind generation has close relationship with both wind speed and wind direction. Especially, the wind power has significant correlation to wind speed, which is consistent with the basic principle of wind energy.

TABLE II. CROSS-CORRELATION COEFFICIENTS BETWEEN WIND GENERATION AND WIND PARAMETERS

| Weather forecast | Cross-correlation coefficients |
|---|---|
| $V$ (10 m) | 0.678 |
| $\theta$ (10 m) | 0.226 |
| $V$ (100 m) | 0.717 |
| $\theta$ (100 m) | 0.219 |

From the correlation analysis, we choose historical wind generation of the latest three hours from the execution time and weather forecast (wind speed and wind direction) at two heights as the input data of the proposed forecast model.

### B. Analysis of MMC Components

The forecasted PDFs with look-ahead time up to 24 hours according to the combined model are displayed in Fig. 4, as well as the forecasted expected value and the actual observations of wind generation. It can be seen that the forecasted PDFs are time-varying and close to center at the observations with different shapes.

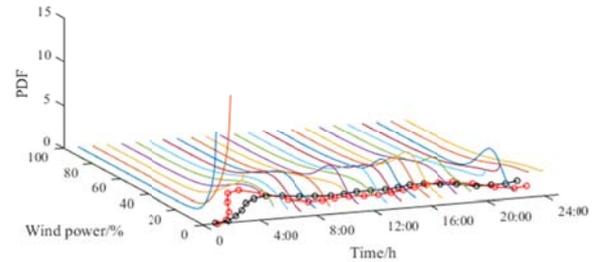

Fig.4 24-hours-ahead predictive densities of the wind generation and the actual observations on 2 June, 2012 (colorful lines: predictive densities; black line with circles: predictive expectations; red line with circles: actual observations).

As an example, the 16-hour-ahead predictive PDF (2 June 2012) obtained by MMC and the corresponding individual model are shown in Fig. 5. The predictive PDF of MMC is a weighted sum of the three individual PDFs. It can be found from Fig. 5 that the observation falls within the 80% central nominal coverage interval of the predictive PDF. The predictive expected value is close to the observation.

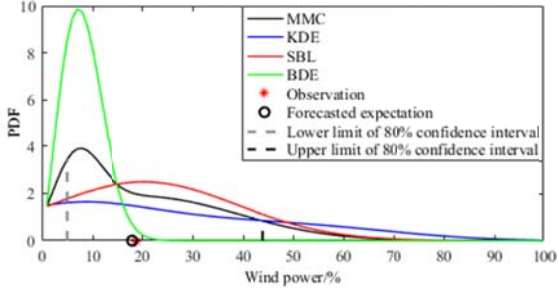

Fig.5 MMC predictive PDF and its three components for the 16-h-ahead wind generation forecast on 2 June, 2012.

The weights $w_k$ of the MMC model reflect the overall contribution of the $k$th combined member. The weights of different look-ahead times and different wind farms are shown in Figs. 6, which demonstrates that weights of the combined components vary with time and space.

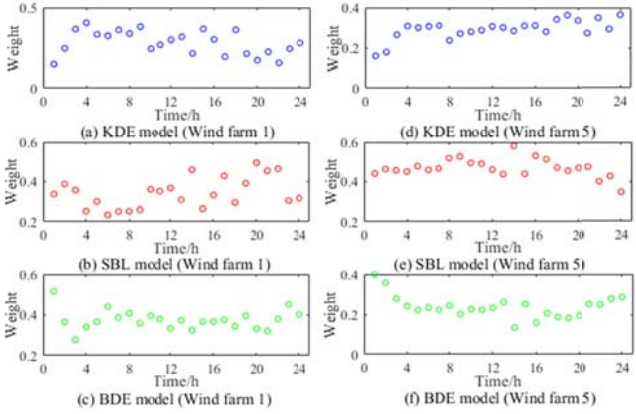

Fig. 6 The weights of three member models of MMC over the evaluation period for wind farm 1 and 5.

### C. Evaluation of Predictive Expectation

Mean absolute error (MAE) and root mean square error (RMSE) are effective indicators to estimate the accuracy of the forecast expectations. The individual probabilistic forecasting models and MMC model with only EM solution are chosen as benchmarks to verify the performance of the proposed MMC model. All the indicator values are the average results of 5 wind farms. The accuracy forecast expectations is shown in Table III. Obviously, the MAE and RMSE values of the MMC model are smaller than the individual models. Moreover the predictive expectations of MMC model are more accurate with further optimization (EM+FO) by CRPS.

TABLE III. ACCURACY OF PREDICTIVE EXPECTATIONS (%)

|      | MMC (EM+FO) | MMC (EM) | KDE   | SBL   | BDE   |
|------|-------------|----------|-------|-------|-------|
| MAE  | 13.32       | 13.44    | 14.45 | 13.74 | 13.88 |
| RMSE | 18.14       | 18.28    | 19.02 | 18.77 | 18.82 |

### D. Evaluation of Predictive PDFs

Reliability and sharpness of the MMC predictive PDFs are evaluated to comprehensively validate its effectiveness. The benchmarks are the same as the above subsection. All the indicator values are also the average results of the 5 wind farms.

#### 1) Reliability

The reliability measures the probabilistic performance of the predictive distributions. It is represented by the bias between the nominal coverage rate of the prediction intervals and the observed coverage rate [33]. Note that the study focuses on central prediction intervals centered at the medium of the probability density function. The bias $R_t^{1-2\alpha}$ is defined as

$$R_t^{1-2\alpha} = \left( \frac{N_t^{1-2\alpha}}{N} - (1-2\alpha) \right) \times 100\%, \quad (29)$$

where $N$ is the size of the test wind power dataset, $N_t^{1-2\alpha}$ is the number of observations covered by the nominal coverage rate $(1-2\alpha)\times 100\%$ of the prediction intervals $[y_\alpha, y_{1-\alpha}]$ and $\Pr(y_\alpha < y < y_{1-\alpha}) = 1-2\alpha$. The closer the nominal coverage rate of the prediction intervals is to the observed coverage rate, the more reliable the prediction intervals are. In particular, $R_t^{1-2\alpha} = 0$ refers to the ideal forecast.

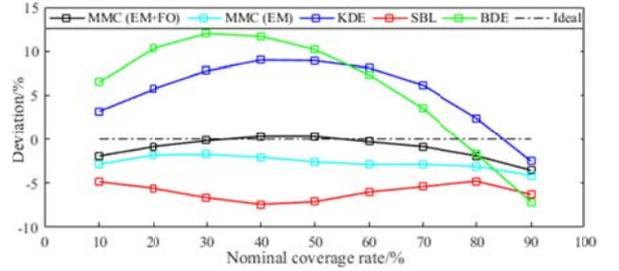

Fig.7 Comparison of different forecasting methods: reliability results with different nominal coverage rates.

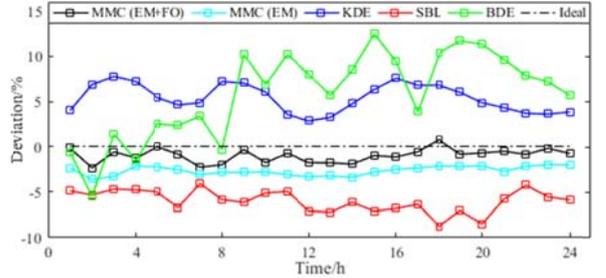

Fig.8 Comparison of different forecasting methods: reliability results with different look-ahead times.

Figs. 7 and 8 show the reliability results of the MMC model and the benchmarks. Fig. 7 displays the average reliability of nominal coverage rates ranging from 10% to 90% with different look-ahead times, while Fig. 8 provides the average reliability results of 24-hours-ahead forecasts with different nominal coverage rates. In Fig. 7, the average absolute deviations of the proposed MMC model is 1.01% and most of them are less than 2%, while those of the SBL, KDE and BDE model are much larger than the MMC, even exceeding 8%. In Fig. 8, most of the absolute deviations fluctuate within the range [0, 2%], much smaller than the member models with different look-ahead times. It is easy to see that the reliability of MMC model is greatly improved with further optimization than with EM only. It is shown that the MMC model can provide more reliable forecasts than the benchmarks.

#### 2) Sharpness

Sharpness metrics evaluate the concentration of predictive intervals. It is known that the sharper the distribution is, the better the uncertainty forecast will be.

For the predictive intervals, the sharpness for the nominal coverage rate of $(1-2\alpha)\times 100\%$ can be measured by

$$S^{1-2\alpha} = \frac{1}{N}\sum_{n=1}^{N}\left|G^{\alpha}(n) - G^{1-\alpha}(n)\right|, \quad (30)$$

where $G^{\alpha}(n)$ and $G^{1-\alpha}(n)$ represent the quantiles of proportions $\alpha$ and $1-\alpha$ of the predictive wind power cumulative density function respectively.

The sharpness results of the MMC model and the benchmarks are shown in Figs. 9 and 10. Fig. 9 depicts the average sharpness results of 10%-90% nominal coverage rates with different look-ahead times, while Fig. 10 displays the average sharpness results of 24-hours-ahead forecasts with different nominal coverage rates. It can be found that the SBL model has sharper interval almost in all the nominal coverage rates, however, it has the poor reliability. In comparisons, the BDE model has the worst sharpness. The interval length of the MMC model with further optimization is a little larger than with only EM. The proposed MMC model has approximately mean value of the sharpness among the models, with interval mean length around 10% smaller than the advanced KDE model.

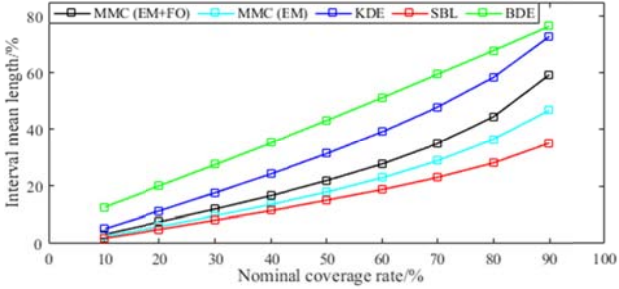

Fig.9 Comparison of different forecasting methods: sharpness results with different nominal coverage rates.

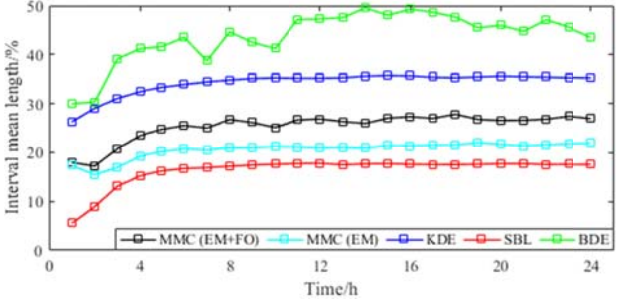

Fig.10 Comparison of different forecasting methods: sharpness results with different look-ahead times.

*3) Comprehensive Evaluation*

As described above, CRPS expressed can be utilized to evaluate the comprehensive performance of the predictive distributions to involve both reliability and sharpness. The CRPS value is larger than zero, and the smaller the better. The CRPS for the forecasting target time *t* can be defined as

$$C_{RPS}(t) = \frac{1}{N}\sum_{n=1}^{N}\int_{0}^{1}\left(F_{t,n}(y) - G_{t,n}(y)\right)^2 dy, \quad (31)$$

where $N$ is the number of forecasting samples, $F_{t,n}(y)$ is the predictive cumulative distribution function (CDF) for the *n*th forecasting sample according to the combined model and $G_{t,n}(y)$ is the experiential cumulative distributions achieved from the observations, described by

$$G_{t,n}(y) = \begin{cases} 0, & y_{t,n}^{*} \leq y \\ 1, & y_{t,n}^{*} > y \end{cases}. \quad (32)$$

The average CRPS values of the studied 5 wind farms for 24 look-ahead hours are shown in Fig. 11. It indicates that the CRPS values of the MMC model are much smaller than the other three individual models except the first three look-ahead times. The priority becomes more obvious with the increasing look-ahead times. The CRPS results demonstrate that the proposed MMC model have more accurate forecasted distributions than the individual models adopted.

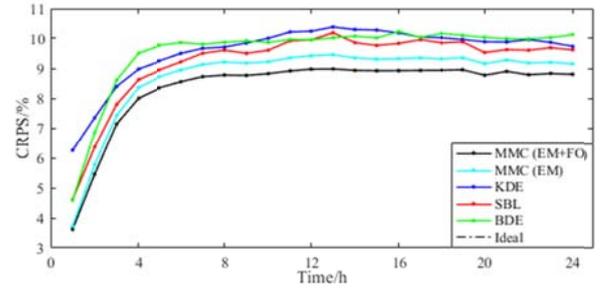

Fig.11 Average CRPS values for 24 look-ahead hours.

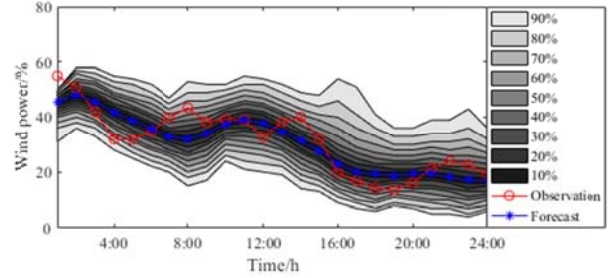

Fig.12 The 24-hours-ahead MMC predictive distributions and actual observations on 6 June, 2012 (Wind farm 2).

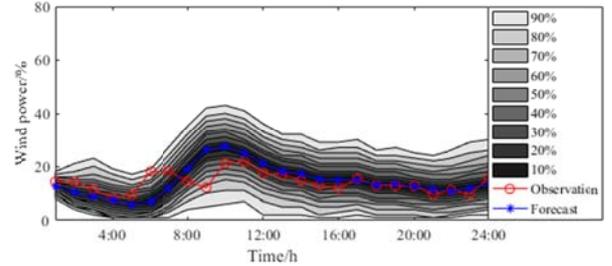

Fig.13 The 24-hours-ahead MMC predictive distributions and actual observations on 2 December, 2012 (Wind farm 2).

The prediction intervals with look-ahead time up to 24 hours obtained by the proposed MMC approach and actual observations in summer (6 June 2016) and winter (2 December 2016) are depicted in Figs. 12 and 13 respectively, where the confidences cover the range 10%-90%. From the examples of Figs. 12 and 13, the predictive intervals can cover the observations pretty well. The predictive distribution significant varies with time, which should be consistent with the heteroscedasticity nature of wind power series.

From the aforementioned numerical results, it can be identified that the reliability and sharpness of the predictive PDFs is opposite, i.e., models with satisfactory reliability usually have poor sharpness, such as the advanced nonparametric model KDE. However, the MMC model with further optimization has better balance than the benchmarks and achieves the best probabilistic forecasting performance. The EM with further optimization can ensure the high quality of predictive distribution of wind power generation. In general, the proposed MMC model can be beneficial for optimal operation and control of power systems with high penetration of wind power.

## V. Conclusions

With the large-scale integration of volatile wind power, accurate wind power forecasting becomes critical to the optimal operation and control of modern power systems. However, it is hard even impossible to obtain precise deterministic forecasting of wind generation. Moreover, it is also difficult to achieve a perfect probabilistic wind generation forecasting model in practice due to the complicated stochastic characteristic of wind power prediction error. In this paper, a novel MMC approach is proposed by extending the conventional BMA model to improve the performance via combining individual forecasting models. The MMC model successfully establishes a weighted combination of several individual probabilistic forecast models conforming to the different distribution forms, both parametric and nonparametric ones. Weights of each member model are firstly estimated by EM and further optimized with respect to the comprehensive performance of the models, assuring its performance in both deterministic and probabilistic forecasts. Numerical study results demonstrate the significant superiority of the proposed MMC model comparing with the adopted individual models.